\documentclass[runningheads]{llncs}
\usepackage{graphicx}
\usepackage{todonotes}

\usepackage{booktabs}

\makeatletter
\newcommand*{\rom}[1]{\expandafter\@slowromancap\romannumeral #1@}
\makeatother

\def\footnoterule{\relax%
  \kern-5pt
  \hbox to \columnwidth{\hfill\vrule width 0.5\columnwidth height 0.4pt\hfill}
  \kern4.6pt}
\makeatother

\begin{document}

\title{Image Segmentation to Identify Safe Landing Zones for Unmanned Aerial Vehicles}

\titlerunning{Safe Landing Zones for UAVs}

\author{Joe Kinahan \and
Alan F. Smeaton\orcidID{0000-0003-1028-8389}}
\authorrunning{J. Kinahan and A.F. Smeaton}
%

\institute{School of Computing and Insight Centre for Data Analytics\\
Dublin City University, Glasnevin, Dublin 9, Ireland.\\
\email{alan.smeaton@DCU.ie}
}

\maketitle

\begin{abstract} 
There is a  marked increase in  delivery services in urban areas, and with Jeff Bezos claiming that 86\% of the orders that Amazon ships weigh less than 5 lbs, the time is ripe for  investigation into economical methods of automating the final stage of the delivery process.
With the advent of semi-autonomous drone delivery services, such as Irish startup `Manna', and Malta's `Skymax', the final step of the delivery journey remains the most difficult to automate. This paper investigates the use of simple images captured by a single RGB camera on a UAV to  distinguish between safe and unsafe landing zones. 
We investigate  semantic image segmentation frameworks as a way to identify safe landing zones and demonstrate the accuracy of lightweight models that minimise the number of sensors needed. By working with images rather than video we reduce  the amount of energy  needed to identify safe landing zones for a drone, without the need for human intervention.  
\end{abstract}
 

\section{Introduction}

Finding appropriate landing sites on land for flying objects in dynamic environments can be used for  both manned and unmanned aerial vehicles. While Irish start-up Manna has shown that there is indeed a market for drone delivery (even securing \$25 million of Series A funding as recently as April 2021), their approach seems to rely on having a human pilot their drones, stating in a press release accompanying their funding announcement that ``a single Manna employee operating multiple drones can operate nearly 20 deliveries per hour".

A large number of factors  affect the suitability of potential landing sites, from rain flooding a site to someone walking their dog in the area. Thus both moving objects and more permanent ground changes can influence the safety of a landing zone.
A well-maintained map of potential landing sites  identified using topological features \cite{kroh2020identification} is a good starting point but using this alone does not lead to sufficient accuracy. 
We focus on the use case of drones being used for deliveries where   GPS and Digital Elevation Models (DEMs) \cite{kakaletsis2019potential} are useful for proposing potential landing sites but live assessment of a proposed site is still necessary. For example, a car park may  appear as an ideal landing site, but what if there is already a car in the spot GPS is directing the drone to or perhaps there is a person walking by as a drone is preparing to land?

Quad-copter drones are the most commonly used type of drone and is  particularly suitable for landing in high density environments as it   needs a landing site only  slightly larger than itself, whereas a fixed wing version requires a `runway' of sorts. 
However, the small size  can  be a disadvantage as such autonomous vehicles require a  range of sensors to compute their position and surroundings thus increasing the weight and space of the device as well as the battery consumption that additional sensors require.

Deploying a trained CNN model on a drone to recognise landing spots can reduce energy consumption by removing the need for an internet connection to send data back and forth to a hub but  we need to ensure that the cost of inference from video sensors is as low as possible. Pruning a trained model is an obvious technique as smaller models  take less time/resources to run, but with video data, we also have other  options. 
First, we can investigate at what frequency we need to process  images -- processing 120 frames per second will be a lot more expensive than 10 but  is there enough of an increase in accuracy to justify the more expensive model and where is the sweet spot? 
Similarly we could reduce the area of an image that we  process.  As a drone drops altitude to land, a camera pointed straight down will cover far more than the area needed  and by cropping  to an  area in the middle of the frame we  reduce the compute cost. 
While it is beneficial to abort a landing as early as possible if it will fail, it is more important to identify obstacles that could actually cause damage to the drone (or the drone could cause damage to) when it is close to the ground. 

The goal of this work is to implement an image-based method to assess the appropriateness of drone landing sites  proposed using topographical features. The novelty of the contribution is in the efficient detection of landing sites allowing on-board computation with reduced energy consumption. In the next section we review current literature on the topic and following that we describe our data collected. That is followed by a section on model training and then a section on video inference, including the model performance in experimental settings. A concluding section completes the paper.

\section{Related Works}

\subsection{Semantic Segmentation}

Convolutional neural networks (CNNs) have shown impressive performance in many  kinds of image segmentation. Results from the  SpaceNet 6 challenge \cite{shermeyer2020spacenet} showed that CNNs are the most effective method for the segmentation of satellite images. The aim of  SpaceNet 6  was to extract building footprints using a Synthetic Aperture Radar (SAR) imagery dataset. SAR is a form of radar that  penetrates clouds and  has the advantage of being able to capture usable data in any weather conditions.
Each of the top-5 performing teams at SpaceNet 6  used an implementation of the U-Net architecture \cite{ronneberger2015u}  which differs from a standard fully connected CNN (FCN)  in that it  is symmetric 
meaning there are the same number of feature maps in the up-sampling path as in the down-sampling path, and the skip connections which skip some layers of the network and feed the output as the input to the next several layers, apply a concatenation operator instead of a sum.
Thus one of its  advantages  is that it does not require all input images to have the same size, allowing cropped areas of images to be classified.

The SpaceNet data identified only  a single class, building footprints, and While buildings are obviously an obstacle to landing aerial vehicles in dynamic settings, there are other objects to consider  such as trees or people. Also, the resolution of the SAR data  at $0.5m^2$ per pixel is likely to be too coarse to be used for identifying drone landing sites. Consequently, it is likely that imagery taken from drones rather than satellite imagery will be more useful in the segmentation we would use. 

While the use case for image classification in  autonomous driving vehicles may  produce a large number of classes like people, cyclists, traffic signals, road markings, other cars and more, aerial imagery typically has low intra-class and high inter-class similarity \cite{bergado2018recurrent}. This suggests that there are broad similarities between different classes of object. Our work    combines several classes in an effort to increase prediction accuracy  as  whether a site is inappropriate because there is a rock or a cyclist in the way does not  matter,  both have the same outcome.

Among the published literature on segmentation, there is disagreement  as to how data augmentation affects the performance of trained models.  In \cite{farabet2012learning} and \cite{ronneberger2015u} the authors credit data augmentations including jitter, re-scaling, rotation and flipping, as a key for learning low-level features. However \cite{long2015fully} reported no noticeable improvement when applying similar data augmentations. Our work  considers the above transformations to see whether they improve accuracy when applied in the UAV Safe Landing Zone (SLZ) context.

The final aspect of our drone landing use case to consider  is  object tracking. As a drone is about to land, there may be moving objects in the vicinity  that cause a proposed site to become unsuitable, such as movement of cars, humans, or animals. There has already been  research in this with regards to autonomous vehicles, often in conjunction with LiDAR \cite{rangesh2019no}. A particularly interesting object tracking framework was  by Xiang {\em et al.} \cite{Xiang_2015_ICCV}, where the object being tracked is modelled with a Markov Decision Process. The object is then classed in one of a number of states - active, inactive, tracked, or lost. Objects in the tracked state  influence whether or not the drone is able to land. If a tracked object  encroaches on a proposed landing site, that site is  rejected in favour of another. An extension of this  could be to model each landing sites as an element of a Markov process where, once identified, the site is placed in either an appropriate or inappropriate state, depending on the presence of  tracked objects.

\subsection{Safe Landing Zone (SLZ) Detection}

There is previous work on automatic detection of SLZs for UAVs. Shah Alam and Oluoch \cite{alam2021survey}  produced a survey of  approaches  split into three main categories --- camera-based, LiDAR-based, and a combination. 
Camera-based approaches  depend on the number of cameras  used by the UAV.  Theodore {\em et al.} \cite{theodore2006flight} proposed a technique using a stereo vision approach where the discrepancy in pixels between the two cameras are used to approximate a terrain profile and a set of constraints is then applied to that  profile to identify appropriate landing sites.

Another camera-based approach is  known as homography estimation where two images of any planar surface are related by a homograhpy, essentially relating the two transformations between two planes. By densely sampling the pixels between successive frames, computing dense optical flow and calculating the homography error, Garg {\em et al.} \cite{garg2018monocular} showed that this could be used to identify whether a proposed landing zone was planar and thus safe to land on, or not. That work focused on the case of water vs. solid ground where  motion on the surface of the water  increases the homography error whereas for solid ground the homography error  is low.

Simultaneous Localisation and Mapping (SLAM) techniques involve using  image sensors on UAVs to create a 3D map of the environment and estimate a UAV's location within that environment. This is particularly useful in unknown environments where there are no landmarks to guide a landing. Yang {\em et al.} \cite{yang2018monocular} proposed a monocular based SLAM technique which was successful in landing a UAV in several different environments. However, it can be difficult to detect SLZs from high altitudes, potentially problematic  because this method involves performing several passes over a larger are in order to map it. This leaves the dilemma of either starting the mapping process too high and getting a poor result, or starting too low and potentially hitting obstacles.

\section{Data Collection}

\begin{figure*}[htb]
    \centering
        \includegraphics[width=0.45\textwidth]{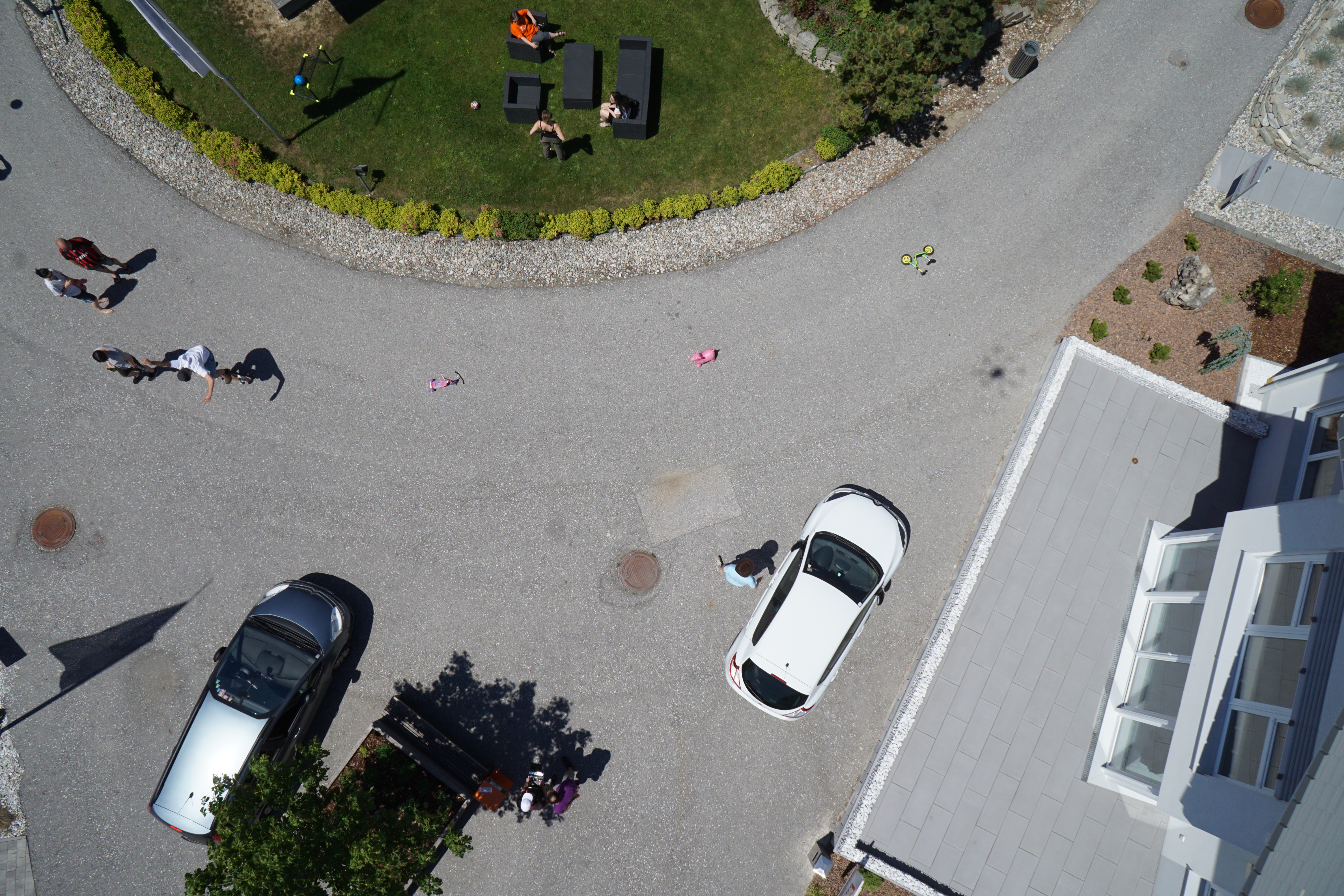}
        \includegraphics[width=0.45\textwidth]{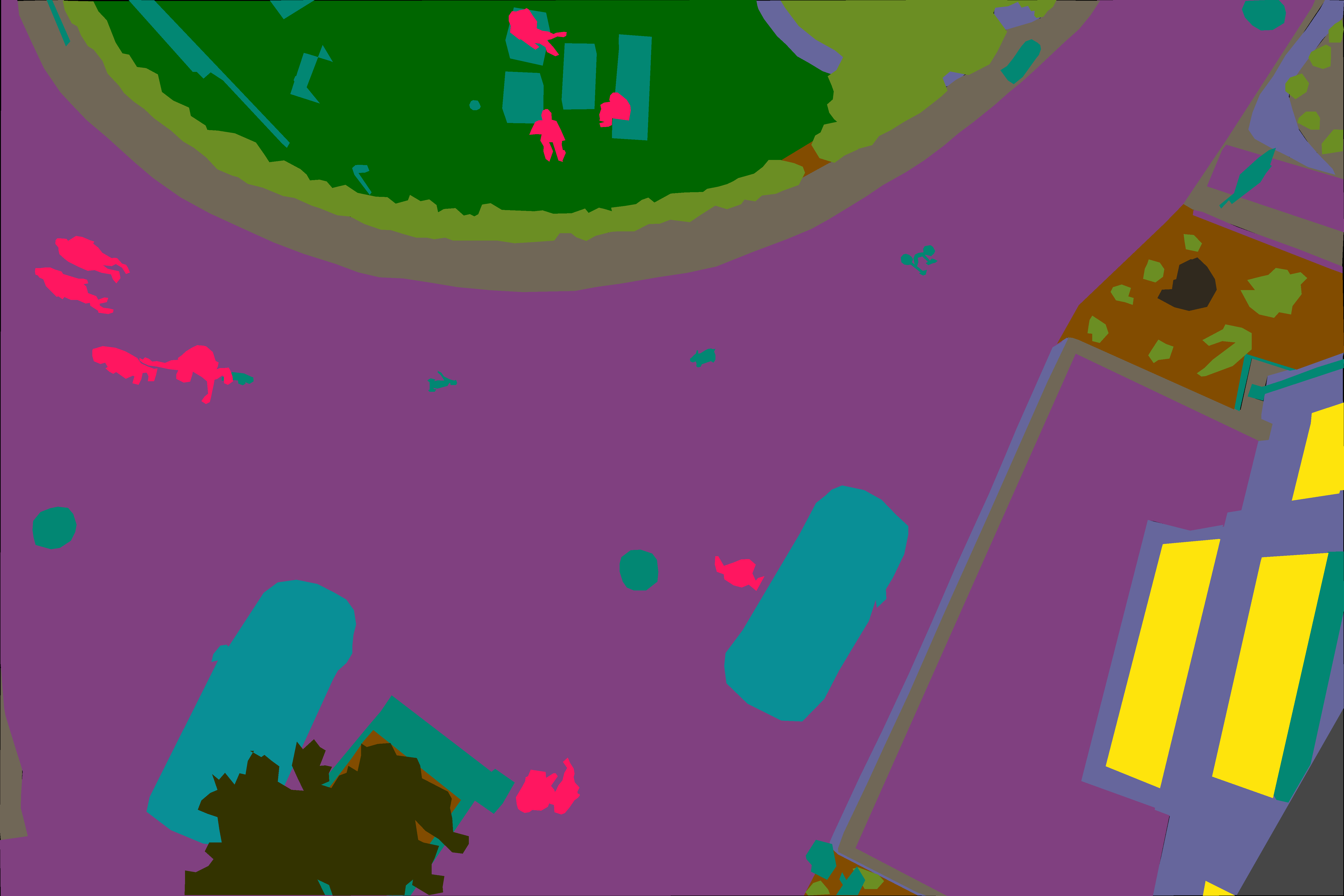}
        \includegraphics[width=0.45\textwidth]{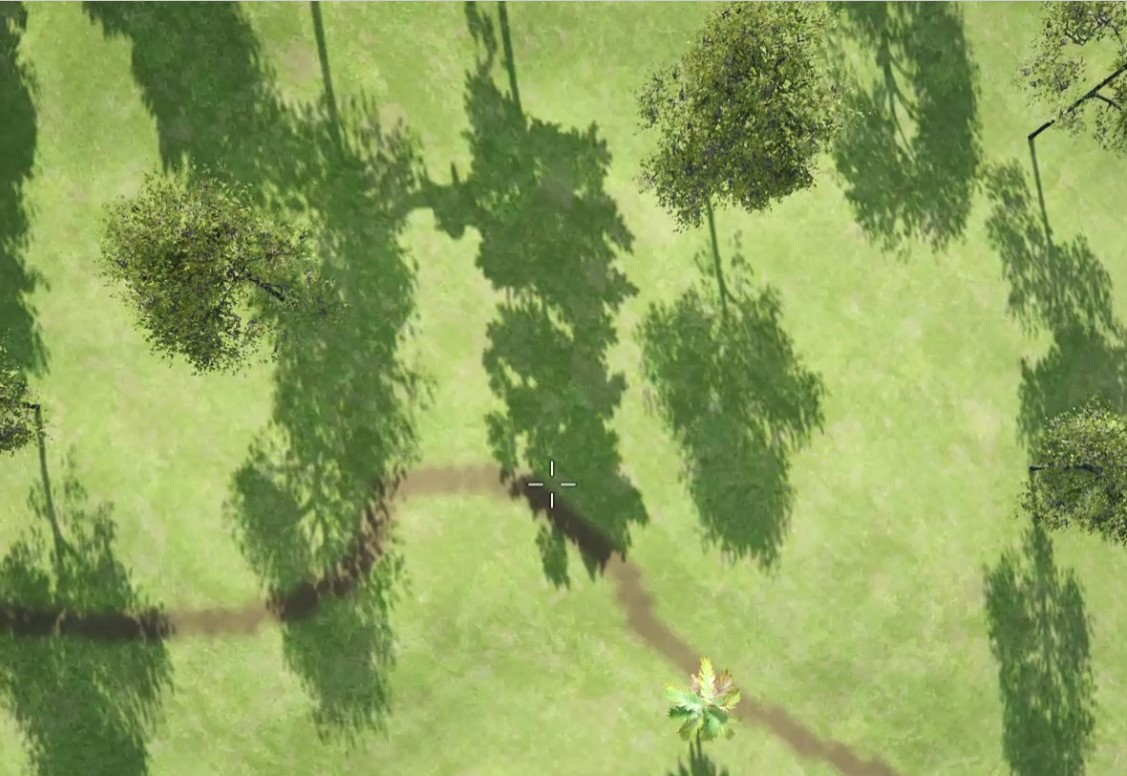}
        \includegraphics[width=0.45\textwidth]{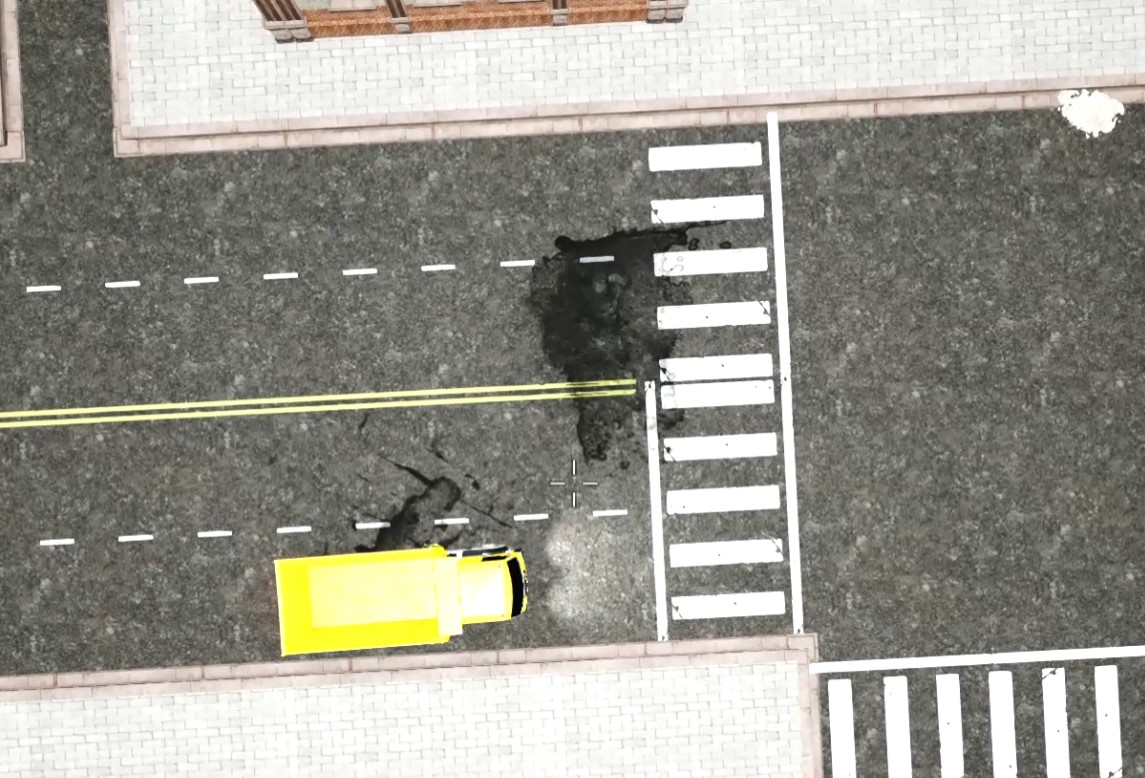}
    \caption{Example aerial photo and its associated mask (top row) and two examples of aerial shots captured in two different environments from the AI Drone Simu game (bottom row) showing how realistic our test videos are.
    \label{fig:sample_photo_mask}
    \label{fig:Drone_env_examples}}
\end{figure*}

Two data sources were used in  work reported in this paper. To train an initial segmentation model, the Aerial Semantic Segmentation Dataset from TU Graz\footnote{http://dronedataset.icg.tugraz.at} was used. This contains 400 images from a nadir (bird's eye) view taken from a height of between 5m and 30m above ground, alongside a mask for each image as shown in Figure~\ref{fig:sample_photo_mask}. The images are  high resolution at 6000*4000px (24 megapixel),   too large for our implementation so the images and associated masks were split into sections and resized to 256*256 pixels. The masks applied were pixel-accurate and  limited to the following classes: tree, grass, other vegetation, dirt, gravel, rocks, water, paved area, pool, person, dog, car, bicycle, roof, wall, fence, fence-pole, window, door, obstacle.
While  these  help provide information about the scene, their primary use is to identify areas that can be landed in namely grass, paved areas, dirt or gravel.

The second  data source we used  were test  videos from drones with downward-facing cameras. For this we simulated drone footage by screen recording a video game. There were several options that could act as drone simulators though few  had a camera that  allowed watching the UAV during flight from a first person perspective and also had the camera  pointing  down at 90\textdegree and a 107\textdegree ~field of view to get the required nadir perspective.
We used a game available on Steam called AI Drone Simu to record  test videos. This  had a number of pre-built environments, some involving moving components such as cars and trains as well as some limited capabilities to create our own environments including cityscapes, empty parks, medieval villages, train yards, and ports. Videos were collected by screen recording while landing the drone. As can be seen in the bottom two images in Figure~\ref{fig:Drone_env_examples}, imagery from this game provided a good and realistic approximation of real life environments. Furthermore, Shafaei {\em et al.} \cite{shafaei2016play} show that video games can reliably be used as a source of training data for computer vision models, and also highlight their advantage in terms of being able to collect data under multiple different environmental parameters.

There were some disadvantages to using the game platform as a simulated environment, mainly   the difficulty of controlling the drone. Whereas in Unity/Blender it would have been possible to define a flight path and speed, in the game environment the drone has to be manually controlled, which can make it difficult to get consistent landings. We disabled any roll, pitch, and yaw motion in the drone and essentially only allowed it to travel on a vertical axis. This still left an issue where the throttle was super sensitive on the controller that was being used and even pressing as lightly as possible often stopped the descent completely. To counteract this, the idle speed of the propellers was increased to the maximum allowed and the drone was allowed to `free-fall' from high enough that it would reach a terminal velocity before landing. The slowest terminal velocity that could be achieved was 15 km/h, which  faster than landing a drone but still allowed us to gather imagery as if the descent rate had been slower.

Another issue was that the game environment does not supply any altitude data. Altimeter sensors are small  and cheap in terms of money and power consumption so it is fair to assume that typically there would be access to  altitude data when flying a drone. Given a fixed speed when landing, we worked backwards from the moment of landing  to approximate the altitude at any given point, computing the descent rate at approximately 10cm per frame at 15 km.h and 30 fps. This  allows us to compare the accuracy of our image segmentation  at different altitudes. 

\begin{figure}[ht]
    \centering
    \includegraphics[width = 0.4\textwidth]{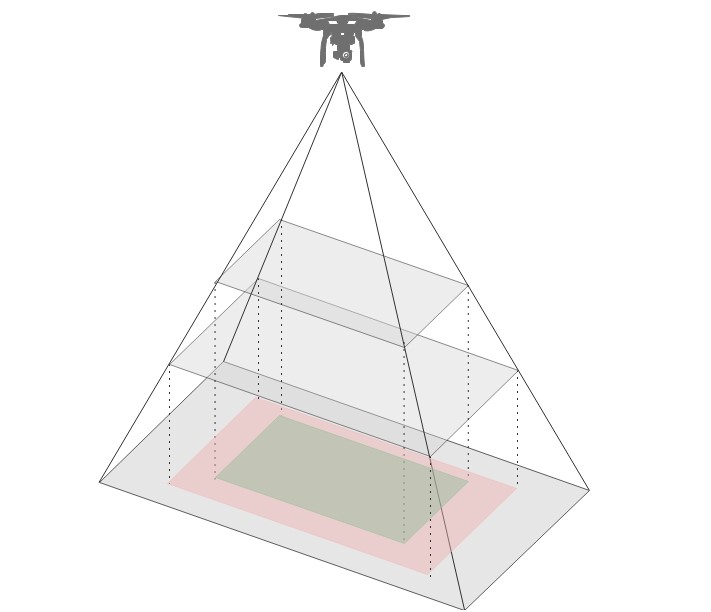}
    \caption{Projection of surface area of subsequent frames onto the initial mask    \label{fig:mask_projection}}
\end{figure}

One of the drawbacks of using this method to collect data is that there is no associated ground truth mask for each of the landing videos. Creating a mask for every frame of 32 videos of 10 seconds duration with each  at 30 fps (more than 9,000 frames in total), is obviously too great. We created a mask for the first frame of each video and that area in subsequent frames was projected back on to this initial frame, which was then cropped to represent the area captured in later frames as shown in Figure~\ref{fig:mask_projection}. 
Here,  the grey areas represent the ground area at frames 1, 2, 3, \ldots, the red area represents the area of frame 2 projected back on to the initial frame, and similarly the green area represents the area of frame 3 projected back to the initial frame. 
This should support a good approximation of the accuracy of  segmentation at different altitudes when in static environments. However, due to the fact that only the initial frame is given a ground truth mask, any dynamic objects would not have a mask associated with them. Thus, the performance of the segmentation is expected to appear worse with  metrics based on this mask for video clips which contain dynamic objects.

\section{Model Training}

As mentioned earlier, the  architecture used in our model is  UNet \cite{ronneberger2015u} with some minor changes  mainly to the input and output layers. Where the original used a single layer input image (in grey scale), here an RGB image was used and  this layer was expanded to three channels. Similarly the output layer,  originally just a binary classification was expanded to a multi-class classification. While ultimately we will use a binary classification (safe or not safe), for the purpose of diagnosing the effectiveness of various models, we will use multi-class and the average of the classes as the value for binary classification.

The classes that were deemed as safe landing zones were grass, gravel, dirt, and paved area. However there were also some `confounding' classes present - areas which look clear and safe to land on, but would ultimately be an inappropriate landing zone. These classes were roofs and water.Thus we consider 3 different models; one containing all safe landing classes, one containing only the dominant safe landing classes, and one containing both the dominant safe landing classes as well as the two confounding classes.

As there are multiple classes being identified, the choice of loss function must reflect that. Because the goal is to classify each individual pixel of the input image, we use a loss function which returns a probability distribution for each pixel and we then take the maximum class probability and assign the pixel to that class. One such loss function is the categorical cross entropy function where an estimate of the probability of each class is made before calculating the entropy between that distribution and the probability distribution of the actual label which will be 1 for the class it belongs to and 0 for everywhere else. Note that using a softmax activation is recommended in the output layer to ensure output remain between 0 and 1
We  used the ADAM optimiser  to find the network weights which minimise this loss function.

To assess the performance of various trained models, the Intersection-Over-Union (IOU), also known as the Jaccard index was used. This  counts the number of pixels which overlap between the predicted mask and the ground truth, divided by the area of union between the two images. Using the five classes mentioned above --  grass, paved area, dirt, gravel, and other -- the mean IOU for the initial model was 0.4762.
While  mean IOU is a good general metric, to see which classes were  classified better than others and perhaps improve it, it is important to look at  individual IOU scores for each class as shown in Table~\ref{tab:Model 1 - Class IOUs}. 

\begin{table}[ht]
    \centering
    \caption{IOU performances for various models where ``$\longrightarrow$'' indicates that class was incorporated into ``Other"
     \label{tab:Model 1 - Class IOUs}
    \label{tab:Model 2 - Class IOUs}
     \label{tab:Model 3 - Class IOUs}}
    \begin{tabular}{l|llllll|l|l}
    \toprule 
 Model~~      & ~Grass~~ &  Paved~~ & Dirt~~ & Gravel~~ &Water~~ & Roof ~~ & ~~Other~~ & ~~Mean~~ \\
\midrule
Model 1~~ & ~0.749 & 0.667 &  0.104 & 0.313& $\longrightarrow$ & $\longrightarrow$ & ~~0.548 & ~~0.477 \\
Model 2 & ~0.719 & 0.653 & $\longrightarrow$ &  $\longrightarrow$ & $\longrightarrow$ & $\longrightarrow$ & ~~0.591 & ~~0.654 \\
Model 3 & ~0.749 & 0.695 & $\longrightarrow$ &  $\longrightarrow$ & 0.220 & 0.448 & ~~0.585 & ~~0.540 \\
\bottomrule 
    \end{tabular}
\end{table}

\noindent 
Interestingly, the initial model was good at classifying both grass and paved areas, but was considerably worse at identifying both gravel and dirt. Examining some of the predictions and their associated masks that while, from the description of gravel and  dirt they sounded like somewhat acceptable landing areas, they were actually unsuitable in a lot of cases. Often they appeared at the edges of other areas such as flowerbeds, trees, fencing etc. 
So based on this a second  model was trained, this time including dirt and gravel as part of  the `other' class. This second model  performed much better as a whole, with an improved mean IOU of  0.654 as shown in Table~\ref{tab:Model 2 - Class IOUs}, however it is again important to check the IOU for each of the individual classes and as we can see, while the mean IOU increased for the second model,  both the grass and paved area classes had poorer accuracy, which are the two that we are most interested in classifying correctly.

Looking at some of the predicted masks vs. their ground truths  helped inform what was happening and we observed that there were two other classes that were frequently being mis-classified as either grass or paved area. Those were water and roof which makes sense since they are each largely flat, unfeatured surfaces. To  improve performance, we  included these as individual classes to be predicted in the third model who's results are also shown in Table~\ref{tab:Model 3 - Class IOUs}. This caused a  decrease in mean IOU, dropping to 0.540 and while this model wasn't particularly good at correctly classifying either water or roof, the improved IOU for the grass and paved area classes as well as the others when compared to the first model, justify this third model as the most appropriate  and it is the one used for inference on the videos.

\section{Video Inference}

In an ideal world, not every frame of incoming video from a UAV should require processing in order to compute a safe landing zone thus reducing energy consumption. We  explored this by examining  how much a predicted mask  in the same area changes during UAV descent. Similar to how the mask on frame 1 was cropped for each frame as shown in Figure~\ref{fig:mask_projection}, we can use the estimate for height and  field of view of the camera to  project the area of the final frame back onto each prior frame and compute how  prediction  changes via the IOU value.

\begin{figure}[ht]
    \centering
    \includegraphics[width = 0.48\textwidth]{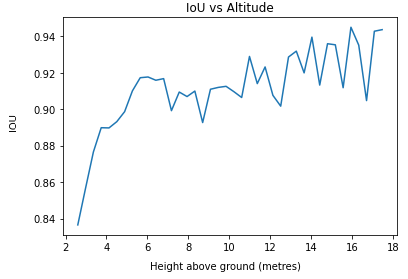}
    \caption{IOU between successive frames vs UAV altitude (in meters)
    \label{fig:iou_vs_altitude}}
\end{figure}

\begin{figure}[ht]
    \centering
    \includegraphics[width = \textwidth]{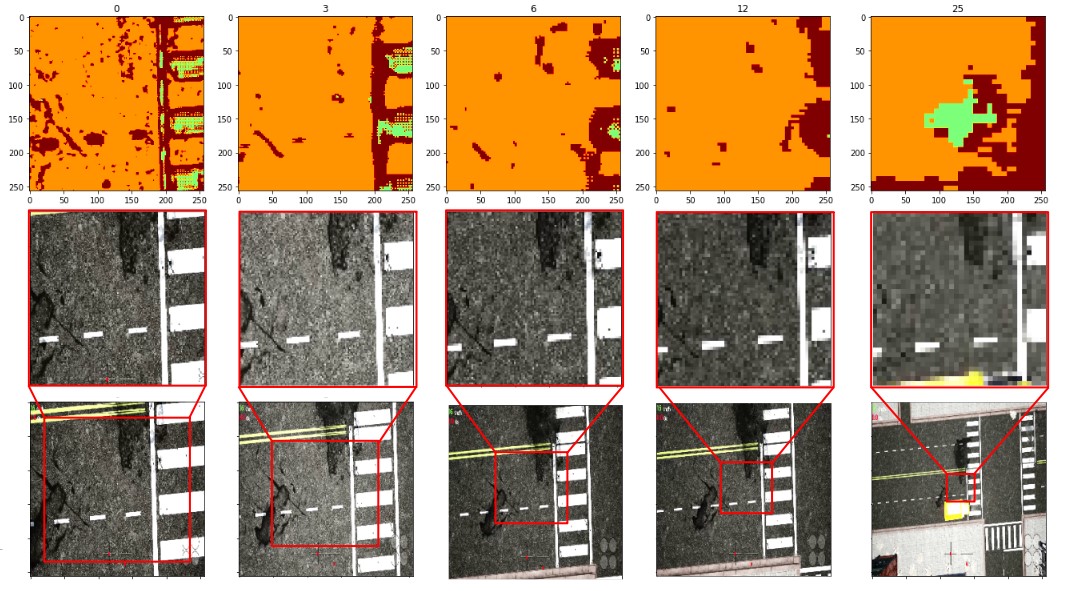}
    \caption{Results of  segmentation of the same surface area from multiple heights
    \label{fig:multiple segments}}
\end{figure}

When we plot  IOU vs. altitude  in Figure~\ref{fig:iou_vs_altitude} we see  that at very low altitudes i.e. the very last few frames of the descent, the predicted masks  vary more between subsequent frames. This is because at low altitudes, the contrast between adjacent pixels of certain surfaces is more exaggerated, whereas at higher altitudes those surfaces  appear  smoother. This is visible in Figure~\ref{fig:multiple segments} where  in the first mask,  pixel-wise prediction is a lot grainier than in subsequent frames. 

This sequence of images and masks also highlights the ability of our model to detect dynamic objects. While not necessarily able to `track' moving objects, by computing the IOU between subsequent frames we can infer whether or not a moving object has entered the frame (or the landing zone), thus potentially making it unsafe to land there. In frame 25 (the rightmost image of Figure~\ref{fig:multiple segments}), it is  possible to make out the edge of what is a truck moving through the picture at the bottom of the image. It can also be seen in the associated mask that there is now an additional area  classed as ``other". This aspect is important to consider when we compare predictions across subsequent frames as dynamic objects move through the frame the segmentations will vary  compared to  previous frames. However, with an average IOU of at least 0.84 at the very low altitude (Figure~\ref{fig:iou_vs_altitude}), there is a  high level of consistency across adjacent images  indicating that the model does not jump between different classifications for large areas of the image.

Given that our model produces predictions which are smooth and consistent even though it works on individual frames sampled from video, we now  check their accuracy using the approach illustrated in Figure~\ref{fig:mask_projection}. In this we start with a ground truth mask manually applied to the first (i.e. highest) frame, the total surface area of the subsequent frame was projected back to the previous frame, cropped from the mask, and becomes the mask of the subsequent frame. 

\begin{figure}[ht]
    \centering
    \includegraphics[width = 0.435\textwidth]{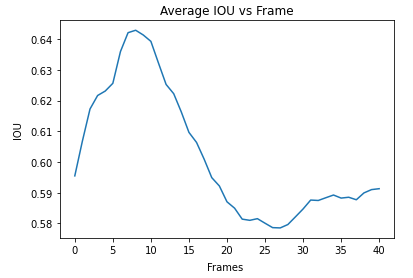}
        \includegraphics[width = 0.555\textwidth]{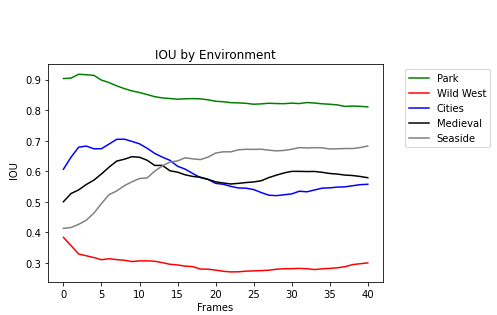}
    \caption{Average IOUs between predicted masks and ground truths at each successive frame across 32 landing videos (left) and average IOU between predicted masks and ground truths at each successive frame for 32 videos in different environments (right)
    \label{fig:iou_ground_truth}
     \label{fig:iou_ground_truth_environment}}
\end{figure}

We note that the mask applied was a simple binary mask with two classes --- safe or unsafe landing zone (SLZ). This  means combining the predicted classes into  groupings, namely combining the paved area and grass classes into safe, and the remainder as unsafe. It also means that the resulting IOUs cannot be directly compared to the IOUs calculated when assessing the earlier models. If we look at the mean of the IOUs across all 32 landing videos in the graph on the left of Figure~\ref{fig:iou_ground_truth} we see that  the classification  is clearly better in the first 15-18 frames which is effectively at heights up to about 10m. 

Examining the averaged IOUs in different landing  videos in the graph on the right of Figure~\ref{fig:iou_ground_truth_environment} gives a clearer picture and shows  a large disparity in the IOUs for different environments. Notably, our model performed  poorly in the `Wild West' environment. Several factors  contributed to the poor segmentation here. The first was the ground cover which appears to be some kind of brushy, dusty surface that looks like it would be reasonably acceptable on which to land a drone. However, that kind of surface did not exist in any of the training images. Another factor may have been the presence of flat roofs throughout the scenes. Much like with the initial training and testing data, our model struggled with classifying these correctly and ended up designating them as safe even though they were not. Finally, some of the `Wild West' scenes contained moving trains and again the roofs of these were reasonably large, unfeatured areas and were often mis-classified. 
On the other hand Our model  performed well in the park scenes which   were deliberately chosen to  contain  potential obstacles for the drone. Consider the park scene in the bottom left in Figure~\ref{fig:Drone_env_examples}, not only does it contain trees (potential obstacles), but also the shadows of the trees which are easy to mis-classify. 

In addition to IOU of particular relevance for evaluating drone landing is the False Positive Rate (FPR), defined as 

$$ FPR = \frac{FP}{FP+TN}$$

\noindent 
Here we  take any unsafe zones that are classified as safe to be a false positive ($FP$), and correctly classified unsafe zones as true negatives ($TN$). 
The value of $FPR$ as a metric is that it is  something that we wish to minimise as there is  a higher cost associated with attempting to land a drone in an unsuitable place than there is in rejecting a suitable one. The former can cause the loss of the UAV, whereas the latter simply means either searching for a new landing site, or aborting the landing.

\begin{figure}[htb]
    \centering
    \includegraphics[width = 0.73\textwidth]{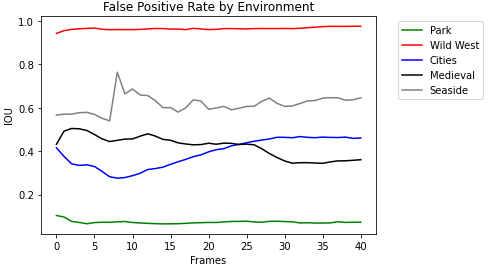}
    \caption{Average False Positive Rate between predicted masks and ground truths at each successive frame for different environments
    \label{fig:fpr_ground_truth_environment}}
\end{figure}

\vspace{-0.25cm}
\noindent 

As shown in Figure~\ref{fig:fpr_ground_truth_environment}, the results for $FPR$ are  similar to IOUs for each environment, performing best in the park scenes, and poorly in the Wild West scenes. Again this poor performance seems to be driven largely by the mis-classification of roofs as safe, which also seems to happen in the city environment. The second poorest environment was the seaside  where, similar to the initial training data, the model struggled to differentiate between  large unfeatured surface areas of bodies of water and unfeatured swathes of land. What was interesting, however, was that for  the seaside environment, the model performed best the closer it was to the ground/surface of the water. This is due to the texture rendering  of the game, where as the drone approached the water, more details  appear such as ripples and small breaking waves and that was a factor in the the sudden jump in false positives between the 5th and 10th frames.

\section{Conclusions}

In this paper we investigated a lightweight, frame based implementation of on-board safe landing zone detection for UAVs. Our implementation was based on exploiting the differences between image segmentation outputs in individual video frames, spaced apart. Our  experiments evaluated the effectiveness of our model on landing videos from downward-facing cameras captured from a gaming platform which used multiple landing environments. Our results show  that there is a lot of potential for lightweight segmentation models to be used to perform real-time on-board segmentation of aerial images and this represents the novel contribution of our paper.

One of the  lessons from our results  is that the performance of  models  varies greatly depending on the landing environment. Unsurprisingly,  environments for which there was  little similarity with any of the training data performed poorly.  However model accuracy  in environments where it had seen similar data in the training suggests that  performance in different environments would improve by  increasing the variety of training data available for those  environments.

\bibliographystyle{splncs04} 
\bibliography{bibfile} 

\begin{thebibliography}{10}
\providecommand{\url}[1]{\texttt{#1}}
\providecommand{\urlprefix}{URL }
\providecommand{\doi}[1]{https://doi.org/#1}

\bibitem{bergado2018recurrent}
Bergado, J.R., Persello, C., Stein, A.: Recurrent multiresolution convolutional
  networks for vhr image classification. IEEE Transactions on Geoscience and
  Remote Sensing  \textbf{56}(11),  6361--6374 (2018)

\bibitem{farabet2012learning}
Farabet, C., Couprie, C., Najman, L., LeCun, Y.: Learning hierarchical features
  for scene labeling. IEEE Transactions on Pattern Analysis and Machine
  Intelligence  \textbf{35}(8),  1915--1929 (2012)

\bibitem{garg2018monocular}
Garg, R., Yang, S., Scherer, S.: {Monocular and Stereo Cues for Landing Zone
  Evaluation for Micro UAVs}. arXiv preprint arXiv:1812.03539  (2018)

\bibitem{kakaletsis2019potential}
Kakaletsis, E., Nikolaidis, N.: Potential {UAV} landing sites detection through
  {Digital Elevation} models analysis. In: European Signal Processing
  Conference, Satellite Workshop (EUSIPCOW) (2019)

\bibitem{kroh2020identification}
Kroh, P.: Identification of landing sites for rescue helicopters in mountains
  with use of {Geographic Information Systems}. J. Mountain Science
  \textbf{17}(2),  261--270 (2020)

\bibitem{long2015fully}
Long, J., Shelhamer, E., Darrell, T.: Fully convolutional networks for semantic
  segmentation. In: Proceedings of the IEEE Conference on Computer Vision and
  Pattern Recognition. pp. 3431--3440 (2015)

\bibitem{rangesh2019no}
Rangesh, A., Trivedi, M.M.: No blind spots: Full-surround multi-object tracking
  for autonomous vehicles using cameras and lidars. IEEE Transactions on
  Intelligent Vehicles  \textbf{4}(4),  588--599 (2019)

\bibitem{ronneberger2015u}
Ronneberger, O., Fischer, P., Brox, T.: U-net: Convolutional networks for
  biomedical image segmentation. In: International Conference on Medical Image
  Computing and Computer-Assisted Intervention. pp. 234--241. Springer (2015)

\bibitem{shafaei2016play}
Shafaei, A., Little, J.J., Schmidt, M.: Play and learn: Using video games to
  train computer vision models. arXiv preprint arXiv:1608.01745  (2016)

\bibitem{alam2021survey}
{Shah Alam}, M., Oluoch, J.: A survey of safe landing zone detection techniques
  for autonomous unmanned aerial vehicles {(UAVs)}. Expert Systems with
  Applications  \textbf{179},  115091 (2021)

\bibitem{shermeyer2020spacenet}
Shermeyer, J., Hogan, D., Brown, J., Van~Etten, A., Weir, N., Pacifici, F.,
  Hansch, R., Bastidas, A., Soenen, S., Bacastow, T., et~al.: {SpaceNet 6:
  Multi-sensor all weather mapping dataset}. In: Proceedings of the IEEE/CVF
  Conference on Computer Vision and Pattern Recognition Workshops. pp. 196--197
  (2020)

\bibitem{theodore2006flight}
Theodore, C., Rowley, D., Ansar, A., Matthies, L., Goldberg, S., Hubbard, D.,
  Whalley, M.: Flight trials of a rotorcraft unmanned aerial vehicle landing
  autonomously at unprepared sites. In: Annual Forum Proceedings-American
  Helicopter Society. vol. 62(2), p.~1250 (2006)

\bibitem{Xiang_2015_ICCV}
Xiang, Y., Alahi, A., Savarese, S.: Learning to track: Online multi-object
  tracking by decision making. In: Proceedings of the IEEE International
  Conference on Computer Vision (ICCV) (2015)

\bibitem{yang2018monocular}
Yang, T., Li, P., Zhang, H., Li, J., Li, Z.: {Monocular vision SLAM-based UAV
  autonomous landing in emergencies and unknown environments}. Electronics
  \textbf{7}(5) (2018)

\end{thebibliography}

\end{document}